\documentclass[conference]{IEEEtran}
\IEEEoverridecommandlockouts
\usepackage{cite}
\usepackage{amsmath,amssymb,amsfonts}
\usepackage{algorithmic}
\usepackage{graphicx}
\usepackage{pgfplots}
\usepackage{float}
\usepackage{textcomp}
\usepackage{xcolor}
\usepackage{url}
\usepackage{booktabs}
\usepackage{algorithm2e}
\usepackage{comment}
\def\BibTeX{{\rm B\kern-.05em{\sc i\kern-.025em b}\kern-.08em
    T\kern-.1667em\lower.7ex\hbox{E}\kern-.125emX}}
\pgfplotsset{compat=1.14}
\usetikzlibrary{arrows}
\begin{document}

\title{Conditional Markov Chain Search for the Generalised Travelling Salesman Problem for Warehouse Order Picking}

\author{\IEEEauthorblockN{1\textsuperscript{st} Olegs Nalivajevs}
\IEEEauthorblockA{\textit{Computer Science and Electric Engineering} \\
\textit{University of Essex}\\
Colchester, UK \\
\url{olegnalivajev@gmail.com}}
\and
\IEEEauthorblockN{2\textsuperscript{nd} Daniel Karapetyan}
\IEEEauthorblockA{\textit{Computer Science and Electric Engineering} \\
\textit{University of Essex}\\
Colchester, UK \\
\url{daniel.karapetyan@gmail.com}}
}

\maketitle

\begin{abstract}
The Generalised Travelling Salesman Problem (GTSP) is a well-known problem that, among other applications, arises in warehouse order picking, where each stock is distributed between several locations -- a typical approach in large modern warehouses.
However, the instances commonly used in the literature have a completely different structure, and the methods are designed with those instances in mind.
In this paper, we give a new pseudo-random instance generator that reflects the warehouse order picking and publish new benchmark testbeds.
We also use the Conditional Markov Chain Search framework to automatically generate new GTSP metaheuristics trained specifically for warehouse order picking.
Finally, we report the computational results of our metaheuristics to enable further competition between solvers.
\end{abstract}


\section{Introduction}
\label{sec:intro}

The Generalised Travelling Salesman Problem (GTSP) is an well-known extension of the Travelling Salesman Problem (TSP)\@.
In GTSP, you are given a set of nodes partitioned into clusters.
You are also given the cost of travelling between each pair of nodes (in this paper we assume that the distances symmetric).
The objective is to find the shortest cycle that visits exactly one node in each cluster.

GTSP has significantly higher modelling power compared to TSP\@.
Many real world applications can be modelled using GTSP with a good accuracy.
Consider any delivery problem; it is common that a delivery driver can choose between several nearby locations (such as either side of the road) where to stop the truck.
Note that these locations might be very distant in terms of driving (think of reversing a truck) and thus this decision may have a significant effect on the cost of the route.
Another example which we will consider in more detail in this paper is the warehouse order picking problem~\cite{IJATES61}.
Assuming that each stock is located in one place within the warehouse, the order picking problem can be modelled using TSP\@.
However, if a stock is distributed between several locations (potentially remote), one needs to use GTSP to model the problem.
Modern warehouses deliberately distribute their stocks between different locations, partly because this allows shorter order picking times.

There are numerous studies of GTSP algorithms.
Some successful heuristics were proposed in~\cite{Karapetyan2012,KARAPETYAN2011221,Helsgaun2015,SMITH20171}.
Particularly effective approaches include metaheuristics such as~\cite{Gutin2009a,Silberholz2007,Pintea2017}.

In this paper, we focus on the warehouse order picking application.
Our contributions are as follows:
\begin{itemize}
    \item We developed a pseudo-random instance generator that produces instances simulating the warehouse order picking.
    We note here that our instances have structure completely different to that commonly used in the literature.
    
    \item Using our instance generator, we produced two testbeds, with medium and large instances.
    
    \item We use the Conditional Markov Chain Search to automatically design a metaheuristic tuned for warehouse picking instances.
    
    \item We give our solutions to our benchmark instances thus enabling other researchers to compare their methods to our solver.
\end{itemize}

\section{Generation of a Metaheuristic}

Conditional Markov Chain Search (CMCS) is a modern framework designed for automated generation of optimisation heuristics~\cite{CMCS-BBQP,KarapetyanParkesStuetzlenst2018,SPLP-CMCS}.
It is a single-point metaheuristic based on multiple components treated as black boxes.
Each component is a subroutine that takes a solution and modifies it according to the internal logic.
Examples of components are hill climbers and mutations.
The behaviour of the control mechanism within CMCS is defined by a set of numeric parameters thus enabling automated generation of CMCS configurations; by tuning these parameters, one can find the `optimal' control mechanism.
Despite being defined by only a small set of numeric parameters, CMCS supports a wide range of control mechanisms.
E.g., it can model Variable Neighbourhood Search, (Weighted) Random Hyperheuristic and Iterated Local Search~\cite{CMCS-BBQP}.

CMCS performs as follows.
It takes as an input the initial solution (usually produced by some construction heuristic, e.g.\ random solution) and then at each iteration applies one of the components to it.
The component modifies the solution according to the internal logic.
The modification is always `accepted', i.e.\ there is no backtracking\footnote{While CMCS control mechanism always accepts any changes, whether improving or worsening the solution, the components such as hill climbers may internally include backtracking.}.
CMCS only records whether the component improved the solution or not.
The choice of the next component depends only on which component was used in the current iteration and whether it improved the solution.
Thus the sequence of applied components is a Markov chain, and the control mechanism can be defined by two transition matrices: one for the case when the solution was improved and another one for the case when the solution was not improved.
The transitions can be probabilistic, however in this research we only consider deterministic transitions, i.e.\ the transition matrices consist of zeros and ones.

Since CMCS may worsen the solution, it keeps track of the best solution found during the search and at the end returns that solution.

\section{Components}

CMCS requires a pool of components to draw from when generating configurations.
Our pool consists of four components that can be found in the literature, see e.g.~\cite{Karapetyan2012}.

Cluster Optimisation (CO) is a component that selects an optimal route given a fixed sequence of clusters.
Such a neighbourhood is exponential in size but it can be explored in polynomial time as this subproblem can be reduced to the shortest path problem.
Thus, Cluster Optimisation is a Very Large Scale Neighbourhood Search.

Insertion Hill Climber (IHC) is a stochastic improvement component.
It randomly picks a node in the solution, randomly picks a new position for it within the solution and then inserts it into this new position.
If the modified solution is not better than the old one, the change is backtracked.
(In fact, we use incremental evaluation and hence the time complexity of IHC is $O(1)$.)

Order Mutation (OM) is a stochastic component that may improve or worsen the solution which is identical to IHC except that it never backtracks any changes.
In other words, it randomly selects a node in the solution, randomly picks a new position for it within the solution and then inserts it into this new position and returns the modified solution.

Vertex Mutation (VM) is another stochastic component that may improve or worsen the solution.
It randomly picks a node within the solution and then replaces it with a randomly picked node from the same cluster.

\bigskip

There have been several data structures used for storing GTSP solutions.
We adopted the data structure proposed in~\cite{Karapetyan2012}.
It separates the ordering in the tour from the vertex selection.
The ordering is stored in a double-linked list.
As the objects in the list are simply the cluster indices from 1 to $m$, the list is represented by only two integer arrays of size $m$.
The double-linked list is particularly convenient as it naturally represent the cyclic tour.
The vertex selection is another integer array of size $m$.
This data structure enables efficient operations on it, elegant implementations of components and compact (cache-efficient) data structures~\cite{Karapetyan2009b}.

\section{Warehouse Instances}

We developed a Warehouse GTSP Instances Generator which more accurately models warehouse order pickup problem compared to the standard approach adopted in the literature, see e.g.~\cite{Gutin2009a}.
Specifically, we did not assume compact clusters with little overlapping; we argue that in a modern warehouse, the locations of stocks of each item are deliberately distributed across the entire warehouse floor.
Indeed, storing the items compactly defeats the purpose of distributing them in multiple locations; with compact storage, chances of a pickup route visiting a remote location for only one or a few items would increase whereas with an even distribution, there is a good chance of some item locations being close to the other fragments of the tour.

Our instance generator takes two parameters: the number of clusters $m$ and the number of nodes $n \ge m$.
First, it generates coordinates of the nodes on a plane, randomly drawing the $x$ and $y$ coordinates from the range $[0, 200]$.
The distances between nodes are computed using Manhattan distance to reflect the typical topology of warehouses.
We form the clusters by placing one node in each cluster and then distributing the rest of the nodes randomly between the clusters.

Our instance generator can be downloaded from \url{https://csee.essex.ac.uk/staff/dkarap/warehousegtsp/Instance.java}.
The instances are in the format of GTSP Instances Library, see \url{https://csee.essex.ac.uk/staff/dkarap/gtsp.html} \cite{Gutin2009a}.

We generated two benchmark testbeds: Medium and Large, 30 instances each.
The Medium instances range from 150 to 202 nodes and 30 to 44 clusters.
The Large instances range from 550 to 602 nodes and 105 to 119 clusters.
The instances can be downloaded from \url{https://csee.essex.ac.uk/staff/dkarap/warehousegtsp/instances.zip}.
Each instance is given a name in the form `$\langle m \rangle \text{wop} \langle n \rangle$'.

\section{Computational Experiments}
\label{sec:experiments}

The algorithm described in this paper has been implemented in Java, and the experiments were conducted on MacBook Pro 15-inch 2017 (4 core Intel Core~i7, 2.9~GHz processor and 16~GB of memory).

First, we needed to generate CMCS configurations.
To generate a configuration, we use a training instance set.
Each configuration is evaluated on each of the training instances.
Specifically, a solution is produced that visits the first node in each cluster, and the order of clusters in the solution is chosen randomly.
Then CMCS is applied to this solution.
The time budget given to CMCS is calculated as follows:
\begin{equation}
t = \alpha nm \,,
\end{equation}
where $\alpha$ is a constant.
We heuristically selected $\alpha = 1.8 \cdot 10^{-5}$ for Medium instances and $\alpha = 3.6 \cdot 10^{-6}$ for Large instances to make sure that each CMCS configuration is given sufficient time to perform at least a few iterations but at the same time not to run for too long (otherwise many configurations would reach optimal solutions and ranking them would become impossible).

We normalised the objective values obtained by different configurations for each instance:
$$
v'_{cI} = \frac{v_{cI} - P_{0}}{P_{50} - P_{0}} \,,
$$
where $v_{cI}$ is the objective value obtained by configuration $c$ on instance $I$, $v'_{cI}$ is the corresponding normalised objective value and $P_i$ is the $i$th percentile in $v_{cI}$ for all configurations $c$ and fixed instance $I$.
The idea to use $P_{50}$ for the upper bound of the normalisation interval is to focus on the high quality solutions and ignore the outliers that otherwise could have a significant effect on the configuration quality metric.
The quality of a configuration is computed as
$$
q_c = \sum_{I} v'_{cI} \,.
$$
The configuration $c$ that minimises $q_c$ is then selected.

With a pool of four components, there are more than quarter of a million configurations.
Testing all these configurations would be impractical.
We follow the idea proposed in~\cite{SPLP-CMCS} and only consider `meaningful' configurations.
We further restrict ourselves to configurations that use exactly three components.
As a result, our set of configurations is reduced to 2972 configurations.

Using the above methodology, we generated two configurations: one being trained on medium instances (which we call Conf1) and one using large instances (which we call Conf2).
The configurations are shown in Figures~\ref{fig:conf1} and \ref{fig:conf2}.
It is interesting to note that the mutation selected in each of these configurations is VM which would be a weak mutation for the typical instances with compact clusters however in the warehouse order picking instances replacing a node with another node from the same cluster may significantly affect the solution.

To enable future competition between solvers, we also report the objective values obtained in our experiments in Tables~\ref{tab:medium} and~\ref{tab:large}.
The `Time, sec' column gives the time budget, the `Best' column reports the best objective value observed in our experiments.
The `Conf1' and `Conf2' columns report the objective values achieved by the corresponding CMCS configurations.
The winning configuration is underlined in each row.
Conf2 outperforms Conf1 on both testbeds but particularly on the Large instances.
However note that training Conf2 was significantly more expensive computationally.
This also demonstrates that the similarity between the training and evaluation instances is important for performance of the generated CMCS configuration.

\tikzset{vertex/.style={circle, draw, thick}}
\tikzset{edge base/.style={->, >=stealth'}}
\tikzset{improved/.style={edge base, blue!#1!white, bend left=20}}
\tikzset{unimproved/.style={edge base, red!#1!white, bend left=20}}
\tikzset{loop improved/.style={edge base, blue!#1!white, loop above, in=70, out=110, looseness=6}}
\tikzset{loop unimproved/.style={edge base, red!#1!white, loop above, in=60, out=120, looseness=7}}


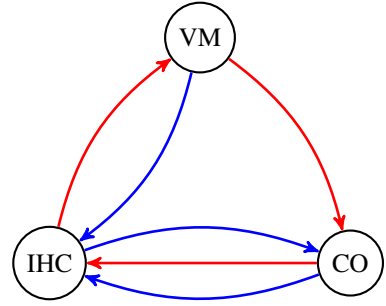
\begin{figure}[htb]
    \begin{center}
	\begin{tikzpicture}
		\useasboundingbox (-2.3,-0.5) rectangle (2.3,3.7);
        
		\node[vertex, align=center] (CO) at (2, 0.0) {CO};
		\node[vertex] (IHC) at (-2, 0.0) {IHC};
		\node[vertex] (VM) at (0, 3.0) {VM};
	    
	    \path (CO) edge[improved=100, line width=1] (IHC);
	    \path (CO) edge[unimproved=100, line width=1, bend left=0] (IHC);
		\path (IHC) edge[improved=100, line width=1] (CO);
		\path (IHC) edge[unimproved=100, line width=1] (VM);
		\path (VM) edge[improved=100, line width=1] (IHC);
		\path (VM) edge[unimproved=100, line width=1] (CO);
   	\end{tikzpicture}
    \end{center}
   	
   	\caption{Conf1, the CMCS configuration trained on small instances.}
   	\label{fig:conf1}
\end{figure}


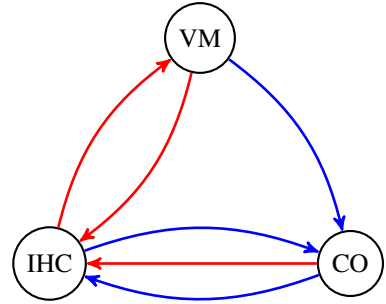
\begin{figure}[htb]
    \begin{center}
	\begin{tikzpicture}
		\useasboundingbox (-2.3,-0.5) rectangle (2.3,3.7);
        
		\node[vertex, align=center] (CO) at (2, 0.0) {CO};
		\node[vertex] (IHC) at (-2, 0.0) {IHC};
		\node[vertex] (VM) at (0, 3.0) {VM};
	    
	    \path (CO) edge[improved=100, line width=1] (IHC);
	    \path (CO) edge[unimproved=100, line width=1, bend left=0] (IHC);
		\path (IHC) edge[improved=100, line width=1] (CO);
		\path (IHC) edge[unimproved=100, line width=1] (VM);
		\path (VM) edge[improved=100, line width=1] (CO);
		\path (VM) edge[unimproved=100, line width=1] (IHC);
   	\end{tikzpicture}
    \end{center}
   	
   	\caption{Conf2, the CMCS configuration trained on medium instances.}
   	\label{fig:conf2}
\end{figure}

\begin{table}[htb]
    \begin{center}
    \begin{tabular}{@{} lrrrr @{}}
    \toprule
    Instance & Best & Time, sec & Conf1 & Conf2 \\
    \midrule
    150wop30&808&0.0810&\underline{948}&980\\
    151wop30&812&0.0815&\underline{826}&1000\\
    153wop31&702&0.0854&\underline{816}&1098\\
    155wop31&724&0.0865&\underline{766}&920\\
    157wop32&694&0.0904&726&\underline{712}\\
    159wop32&774&0.0916&\underline{774}&940\\
    160wop33&876&0.0950&1020&\underline{1002}\\
    162wop33&804&0.0962&1014&\underline{890}\\
    164wop34&914&0.1004&\underline{920}&1068\\
    166wop34&844&0.1016&\underline{898}&1022\\
    168wop35&974&0.1058&\underline{1112}&1236\\
    169wop35&1014&0.1065&\underline{1164}&1166\\
    171wop36&898&0.1108&\underline{898}&1360\\
    173wop36&866&0.1121&1116&\underline{942}\\
    175wop37&884&0.1166&\underline{1050}&1110\\
    177wop37&840&0.1179&1034&\underline{1012}\\
    178wop38&988&0.1218&1058&\underline{1028}\\
    180wop38&1080&0.1231&1282&\underline{1192}\\
    182wop39&978&0.1278&1238&\underline{1048}\\
    184wop39&1084&0.1292&1358&\underline{1238}\\
    186wop40&1032&0.1339&1400&\underline{1264}\\
    187wop40&994&0.1346&1322&\underline{1218}\\
    189wop41&1030&0.1395&1122&\underline{1052}\\
    191wop41&1020&0.1410&1184&\underline{1086}\\
    193wop42&1040&0.1459&1438&\underline{1040}\\
    195wop42&1038&0.1474&1396&\underline{1126}\\
    196wop43&1072&0.1517&1286&\underline{1106}\\
    198wop43&1150&0.1533&1278&\underline{1242}\\
    200wop44&1166&0.1584&1330&\underline{1208}\\
    202wop44&1194&0.1600&\underline{1302}&1382\\
    \bottomrule
    \end{tabular}
    \end{center}
    
    \caption{Computational results for the Medium Warehouse Order Picking instances.}
    \label{tab:medium}
\end{table}

\begin{table}[htb]
    \begin{center}
    \begin{tabular}{@{} lrlrr @{}}
    \toprule
    Instance & Best & Time, sec & Conf1 & Conf2 \\
    \midrule
    550wop105&2958&0.2079&3334&\underline{3306}\\
    551wop105&3170&0.2083&3552&\underline{3170}\\
    553wop106&2288&0.2110&2812&\underline{2698}\\
    555wop106&2224&0.2118&2530&\underline{2224}\\
    557wop107&2432&0.2146&2952&\underline{2534}\\
    559wop107&2264&0.2153&2626&\underline{2348}\\
    560wop108&2444&0.2177&2632&\underline{2544}\\
    562wop108&2486&0.2185&3036&\underline{2686}\\
    564wop109&2270&0.2213&2750&\underline{2270}\\
    566wop109&2402&0.2221&2990&\underline{2574}\\
    568wop110&2430&0.2249&2944&\underline{2670}\\
    569wop110&2246&0.2253&2800&\underline{2692}\\
    571wop111&2246&0.2282&2862&\underline{2246}\\
    573wop111&2338&0.2290&3052&\underline{2500}\\
    575wop112&2358&0.2318&2892&\underline{2466}\\
    577wop112&2444&0.2326&2978&\underline{2762}\\
    578wop113&2296&0.2351&2866&\underline{2342}\\
    580wop113&2600&0.2359&3086&\underline{2626}\\
    582wop114&2518&0.2389&3130&\underline{2584}\\
    584wop114&2188&0.2397&2560&\underline{2188}\\
    586wop115&2588&0.2426&3192&\underline{2588}\\
    587wop115&2612&0.2430&3312&\underline{2802}\\
    589wop116&2650&0.2460&2946&\underline{2650}\\
    591wop116&2600&0.2468&3070&\underline{2696}\\
    593wop117&2584&0.2498&2850&\underline{2696}\\
    595wop117&2450&0.2506&2834&\underline{2450}\\
    596wop118&2748&0.2532&3450&\underline{2974}\\
    598wop118&2384&0.2540&3256&\underline{2554}\\
    600wop119&2574&0.2570&3204&\underline{2574}\\
    602wop119&2548&0.2579&3042&\underline{2548}\\
    \bottomrule
    \end{tabular}
    \end{center}
    
    \caption{Computational results for the Large Warehouse Order Picking instances.}
    \label{tab:large}
\end{table}

We also conducted preliminary tests of Conf1 and Conf2 on the standard instances from the GTSP Instance Library to compare them to the state-of-the-art solvers.
Our early conclusions are that Conf1 and Conf2 outperform the state-of-the-art metaheuristics in terms of the running time on small instances, however perform poorly on larger instances.
This was expected as (a) Conf1 and Conf2 are trained on Warehouse Order Picking Instances and hence are not supposed to perform well on the instances from the GTSP Instances Library, and (b) CMCS is a single-point metaheuristic whereas most powerful metaheuristics are usually population-based.
Further experiments are required to compare CMCS configurations to the other solvers on the Warehouse Order Picking Instances.
We expect to see that the generated CMCS configurations will perform better than the other solvers as they were specifically trained for these instances.
That will be a valuable contribution, as it will demonstrate the advantage of using automated metaheuristic generation.

\section{Conclusions}

This paper discusses an important application of the GTSP to the warehouse order picking.  
We argue that the GTSP benchmark instances commonly used in the literature do not adequately reflect the structure of the warehouse order picking problem and thus the available GTSP solvers might not be well-suited for the warehouse picking problem.
In this paper, we give a new instance generator, benchmark testbeds and automatically generate a metaheuristic using the Conditional Markov Chain Search specifically for the warehouse order picking application.
We then report our computational results.
This is still work in progress, and further experiments will be needed to establish how our methods compare to the existing solvers.

\bibliographystyle{IEEEtran}
\bibliography{refs}

\begin{thebibliography}{10}
\providecommand{\url}[1]{#1}
\csname url@samestyle\endcsname
\providecommand{\newblock}{\relax}
\providecommand{\bibinfo}[2]{#2}
\providecommand{\BIBentrySTDinterwordspacing}{\spaceskip=0pt\relax}
\providecommand{\BIBentryALTinterwordstretchfactor}{4}
\providecommand{\BIBentryALTinterwordspacing}{\spaceskip=\fontdimen2\font plus
\BIBentryALTinterwordstretchfactor\fontdimen3\font minus
  \fontdimen4\font\relax}
\providecommand{\BIBforeignlanguage}[2]{{%
\expandafter\ifx\csname l@#1\endcsname\relax
\typeout{** WARNING: IEEEtran.bst: No hyphenation pattern has been}%
\typeout{** loaded for the language `#1'. Using the pattern for}%
\typeout{** the default language instead.}%
\else
\language=\csname l@#1\endcsname
\fi
#2}}
\providecommand{\BIBdecl}{\relax}
\BIBdecl

\bibitem{IJATES61}
J.~Karasek, ``An overview of warehouse optimization,'' \emph{International
  Journal of Advances in Telecommunications, Electrotechnics, Signals and
  Systems}, vol.~2, no.~3, pp. 111--117, 2013.

\bibitem{Karapetyan2012}
D.~Karapetyan and G.~Gutin, ``Efficient local search algorithms for known and
  new neighborhoods for the generalized traveling salesman problem,''
  \emph{European Journal of Operational Research}, vol. 219, pp. 234--251,
  2012.

\bibitem{KARAPETYAN2011221}
------, ``{Lin-Kernighan} heuristic adaptations for the generalized traveling
  salesman problem,'' \emph{European Journal of Operational Research}, vol.
  208, no.~3, pp. 221--232, 2011.

\bibitem{Helsgaun2015}
K.~Helsgaun, ``Solving the equality generalized traveling salesman problem
  using the lin--kernighan--helsgaun algorithm,'' \emph{Mathematical
  Programming Computation}, vol.~7, no.~3, pp. 269--287, Sep 2015.

\bibitem{SMITH20171}
S.~L. Smith and F.~Imeson, ``Glns: An effective large neighborhood search
  heuristic for the generalized traveling salesman problem,'' \emph{Computers
  \& Operations Research}, vol.~87, pp. 1--19, 2017.

\bibitem{Gutin2009a}
G.~Gutin and D.~Karapetyan, ``A memetic algorithm for the generalized traveling
  salesman problem,'' \emph{Natural Computing}, vol.~9, no.~1, pp. 47--60,
  2009.

\bibitem{Silberholz2007}
J.~Silberholz and B.~Golden, \emph{The Generalized Traveling Salesman Problem:
  A New Genetic Algorithm Approach}.\hskip 1em plus 0.5em minus 0.4em\relax
  Boston, MA: Springer US, 2007, pp. 165--181.

\bibitem{Pintea2017}
C.-M. Pintea, P.~C. Pop, and C.~Chira, ``The generalized traveling salesman
  problem solved with ant algorithms,'' \emph{Complex Adaptive Systems
  Modeling}, vol.~5, no.~1, p.~8, Aug 2017.

\bibitem{CMCS-BBQP}
D.~Karapetyan, A.~P. Punnen, and A.~J. Parkes, ``Markov chain methods for the
  bipartite boolean quadratic programming problem,'' \emph{European Journal of
  Operational Research}, vol. 260, no.~2, pp. 494--506, 2017.

\bibitem{KarapetyanParkesStuetzlenst2018}
D.~Karapetyan, A.~J. Parkes, and T.~St\"utzle, ``Algorithm configuration:
  Learning policies for the quick termination of poor performers,'' in
  \emph{Proceedings of LION 2018}, ser. LNCS, vol. 11353, 2018, pp. 220--224.

\bibitem{SPLP-CMCS}
D.~Karapetyan and B.~Goldengorin, ``Conditional markov chain search for the
  simple plant location problem improves upper bounds on twelve korkel-ghosh
  instances,'' in \emph{Optimization Problems in Graph Theory}, B.~Goldengorin,
  Ed.\hskip 1em plus 0.5em minus 0.4em\relax Springer, 2018, pp. 123--147.

\bibitem{Karapetyan2009b}
D.~Karapetyan, G.~Gutin, and B.~Goldengorin, ``Empirical evaluation of
  construction heuristics for the multidimensional assignment problem,'' in
  \emph{London Algorithmics 2008: Theory and Practice}, ser. Texts in
  algorithmics.\hskip 1em plus 0.5em minus 0.4em\relax London, UK: College
  Publications, 2009, pp. 107--122.

\end{thebibliography}

\end{document}